\documentclass[10pt]{article}

\usepackage{amsmath}
\usepackage{amssymb}

\usepackage{graphicx}

\usepackage{cite}

\usepackage{color} 

\usepackage[labelfont=bf,labelsep=period,justification=raggedright]{caption}

\makeatletter
\renewcommand{\@biblabel}[1]{\quad#1.}
\makeatother

\date{}

\begin{document}

\begin{flushleft}
{\Large
\textbf{Bipartite ranking algorithm for classification and survival analysis}
}
\\
Marina Sapir, 
\\
 MetaPattern, Bar Harbor, ME
\\
$\ast$ E-mail: m@sapir.us
\end{flushleft}

\begin{abstract}
Unsupervised aggregation of independently built univariate predictors is explored as an alternative regularization approach for noisy, sparse datasets. Bipartite ranking algorithm Smooth Rank implementing this approach is introduced. The advantages of this algorithm are demonstrated on two types of problems. First, Smooth Rank is applied to two-class problems from bio-medical field, where ranking is often preferable to classification. In comparison against  SVMs with radial and linear kernels, Smooth Rank had the best performance on 8 out of 12 benchmark benchmarks. The second area of application is survival analysis, which is reduced here to bipartite ranking in a way which  allows one to use commonly accepted measures of methods performance. In comparison of Smooth Rank with Cox PH regression and CoxPath methods, Smooth Rank proved to  be the best on 9 out of  10 benchmark datasets. 
\end{abstract}

\section*{Introduction}

Most of efforts in ``classical'' machine learning were concentrated on classification problems, where number of features is fixed, number of instances can be increased, classes can be smoothly separated in attribute space.  Real life biological problems are  different in every aspect: number of features can be increased almost without limit, new cases are very hard to get, observed outcome is influenced to a large degree by unknown factors. 

The main direction of adaptation of machine learning methods to this reality is regularization. The goal of the regularization is to improve  generalization by putting additional price on smoothness of the model, to make modeling less sensitive to the peculiarities of sparse datasets, high dimensionality and outliers. Most of time the regularization implies restrictions on weights in the model, and includes some tuning hyper-parameters. However, finding the optimal values for these additional parameters on the limited data may further increase method's sample-sensitivity, contrary to the goal of the regularization \cite{ModelSelectionBias, AdaBoostOverfit}.  We confirm this finding in our experiments. 

The approach to regularization, explored here, is based on unsupervised aggregation of univariate predictors. The idea is that when there is not enough data to build a reliable multidimensional model, there may be sufficient data to build independently and aggregate univariate models. Within this approach,  empirical formulas are used for tuning to a particular data to avoid optimization of the hyper-parameters.   

The  direction is not new in machine learning. It generalizes such methods as Naive Bayes and weighted voting \cite{Golub}, which are known for good performance on difficult data \cite{Friedman, HTF, Segal} despite its ``naivity''. In particular, J. Friedman \cite{Friedman} suggested  that success of the Naive Bayes can be explained by its smoothing properties. Here these ideas are implemented  in the new bipartite ranking method Smooth Rank. Effectiveness of Smooth Rank comparing with SVMs with linear and radial kernels and Rank Boost is demonstrated on 12 datasets from bio-medical field, where ranking is often preferable to classification. 

As another contributions of this work,  we show that survival analysis can be effectively reduced to a bipartite ranking problem as well,  and successfully apply Smooth Rank to this difficult problem. In comparison with leading survival analysis methods, Smooth Rank shows the best performance in 9 out of 10 datasets in these experiments. 

The article is organized as follows. In the Section 1,  the bipartite ranking algorithm Smooth Rank is defined and analyzed. In the Section 2, we present and discuss results of experiments on 12 benchmark two-class datasets. In the next section, interpretation of survival analysis as bipartite ranking problem is introduced. In the fourth section, the performance of Smooth Rank is compared with other methods on survival analysis datasets.

\section{Smooth Rank}

\subsection{Definition of Smooth Rank}

The main scheme  of the algorithm can be described as two steps procedure:

\begin{enumerate}
\item Independently for each feature $x^i:$ build a predictor $f_i(x^i)$ and calculate its weight $w_i$;
\item Calculate a scoring function $F(x) = \sum_1^n w_i \cdot f_i(x^i).$

\end{enumerate}

For  a classification problem, there are two popular ensemble algorithms which follow this scheme. One of them is Naive Bayes classifier \cite{HTF} where all weights     $w_i \equiv 1$ and  each predictor  is built as a log-ratio between densities  of two classes:  $f_i(x^i) = log(d_{1, i}(x^i) / d_{-1,i}(x^i)).$  Another example of an algorithm with the same scheme is ``weighted voting''  \cite{Golub}. 

There are several ways the general scheme  can be implemented. Below is one of the possible implementations.\\ 

\textbf{Algorithm Smooth Rank}
Given is sample $\{X, Y\}$, where the response variable $Y$ has two values, 1 and 2.   \begin{enumerate}

\item For each feature $x^i$:
\begin{enumerate}
\item Build kernel approximations $g^i_{1}, g^i_2$ of the density  of each class on $R^i = dom(x^i)$;
\item For each point $r \in R^i$ calculate $$q_i(r) = \frac{ g^i_1(r)  -   g^i_{2}(r) }{\pi_1 \cdot g^i_1(r)  + \pi_2 \cdot g^i_{2}(r)}, $$ where $\pi_1, \pi_2$ are frequencies of the classes 1 and 2;
\item Build marginal predictors  $\widetilde{q_i}(x)$  as LOESS approximation of the function $q_i(x)$
\item Calculate weights $w_i$ of the predictors based on their correlation with outcome
\end{enumerate}
\item Calculate scoring function
$$ F(x) = \sum_{i: \, x^i \neq NA} w_i \cdot \widetilde{q_i}(x^i) / \sum_{i: \, x^i \neq NA}  w_i $$
\end{enumerate}

\subsection{Implementation and parameters}
\subsubsection{Density evaluation}
To improve generalization ability, the evaluation of density of each class is implemented as a two step procedure: first, the density is approximated with cosine kernel (R function \textit{density} ); then the density function is smoothed using LOESS procedure. 

To avoid ``model selection bias'' \cite{ModelSelection, ModelSelectionBias}, we fixed the parameters of these R functions in all our experiments:  \textit{density} was evaluated on equally spaced default 512 points; the procedure \textit{loess} was used with polynomials of the  degree 1.

It is well known that comparison of densities of two classes is not reliable in the intervals of low density \cite{HTF}. To deal with this issue, the function $q(r)$ is not evaluated for  $r: \pi_1 \cdot g^i_1(r)  + \pi_2 \cdot g^i_{2}(r) < 0.1$. The aggregation on the next step handles values of the predictors in these points as missing.

\subsubsection{Calculation of weights and post-filtering}
The calculation of weights is implemented as a two step procedure. 

First, weights of the predictors are calculated by the formula $$w_i = AUC(\widetilde{q_i}(x), Y) - 0.5.$$ 

The next step includes ``post-filtering'', or ``shrinkage''. The goal of  this step is to improve learning on the datasets where  many correlated weak predictors combined would overweight  few strong ones. Rather than setting a hard threshold for selection of predictors, or use data to optimize the threshold,  the filtering is made based on comparison with the best available predictor. The updated weights are calculated by the formula:

$$ w_i := \left\{ \begin{array}{ll}
                   w_i, & \,if \, \, w_i > (\max_j w_j) / 3 \\
                   0, & otherwise.
                   \end{array}
                   \right.
  $$
The empiric  formula allows to filter out relatively weak predictors, making the filtering data-dependent without additional optimization procedure. 

\subsection{Properties of the predictors}

Since $g^i_{1}, g^i_2$ approximate densities of both classes, according to Bayes theorem,   $$ P(Y = j | x^i = r) \simeq \frac{\pi_j \cdot g^i_j(r) }{\pi_1 \cdot g^i_1(r)  + \pi_2 \cdot g^i_{2}(r)},$$ $j = 1,2$. Then the marginal predictor  $\widetilde{q_i}(r) $ can be presented as
$$\widetilde{q_i}(r) \simeq \frac{P(Y = 1 | x^i = r)}{\pi_1} - \frac{P(Y = 2 | x^i = r)}{\pi_2}.$$

If variable $x^i$ is conditionally independent of  $Y$ in $x^i = r$, the conditional posterior probabilities of both classes in this point are equal to their priors,   and $\widetilde{q_i}(r) \simeq 0.$  In each point $r$, the value of marginal predictor function $\widetilde{q_i}(r) $ indicates degree and direction of  local association between the values of the variable $x^i$ and the response variable.   Then, $\widetilde{q_i}(r) $ influences the scoring function $F$ only for those points $x^i = r$  which are predictive, and do not participate in ranking otherwise.

\section{Experiments on  Bipartite Ranking Data}

Bipartite ranking problems can be approached with at least two types of methods: by the ranking methods (as RankBoost \cite{RankBoost}, Rank SVM \cite{RankSVM}), and by any classification method which produces a continuous output (as probabilities) along with binary classification. Based on these considerations,  SVMs with linear and radial kernels and RankBoost were selected for comparison with Smooth Rank on bipartite data.  Function {\it svm} from R package {\it e1071}  was used with default parameters for both types of kernels. Rank Boost was implemented with weak learners as binary functions.  RankBoost was applied with the number of iterations set to 80: preliminary experiments demonstrated that further increasing this parameter does not improve learning. 

Twelve of  two-class biomedical benchmark datasets from UCI Machine Learning Repository were used for testing. Biomedical area was selected because poorly separable classes, typical for such data,   make bipartite ranking especially useful. 

The datasets Cleveland HD, VA HD, Switzerland HD, Hungrian HD represents samples from the same collection, Heard Disease. Even though all the datasets have 13 independent variables, some variables have most of values missing. In each sample, the features with more than 20 $\%$ of missing values were excluded.

\begin{table*}

\caption{Comparison of Bipartite Ranking Methods}
\centering
\begin{tabular}{|l |l|l|l|l|l|l|}
\hline
$\#$ & Data &  Dimensions &  Smooth Rank & SVM radial & SVM linear  & RankBoost\\ 
\hline
1 & Hepatitis  & 155 X 9 
	& 0.86 (12.7)  
	&  \textbf{0.88} 
	& 0.80  
	& 0.80\\

2 & Cleveland HD &  283 X 13 
	& \textbf{0.90} (10.3)
	& 0.88  
	& 0.87
	&  \textbf{0.90} \\
	
3 & Indians & 768 X 8 
	& \textbf{0.83} (6.6)  
	& 0.81 
	& 0.82  
	& \textbf{0.83}\\ 

4 & Statlog-Heart &  270 X 13 
		   & \textbf{0.90} (10.4) 
	       & 0.89  
	       & 0.88 
	       & \textbf{0.90}\\

5 & Parkinsons   & 195 X 22
	& 0.88 (21.7)  
	&  0.86
	& 0.81 
	& \textbf{0.90}\\
	
6 & LjublianaBrCa &  286 X 9 
	& \textbf{0.71} (6) 
	& 0.68
	& 0.66
	& \textbf{0.71}\\
	
7 & Hungarian HD &  294 X 10
	& \textbf{0.88}   (6)
	& 0.86 
	& 0.87 
	& 0.87\\
		
8 & Switzerland HD  &  123 X 8 
	& 0.60 (5)
	&  \textbf{0.61} 
	&  \textbf{0.61}
	& 0.59 \\
	
9 & VA HD &  200 X 6 
	& 0.64 (3) 
	&  \textbf{0.69 }
	& 0.66
	& 0.66\\
	
10 & Wpbc & 198 X 33 
	& 0.67 (17.4) 
	& 0.7  
	& \textbf{0.74} 
	&0.63\\
	
11 & Spine (Column)  &  310 X 6
	& 0.86 (6)
	&  \textbf{0.91} 
	&  0.89
	& 0.71\\

12 & SPECTF   & 267 X 44
	& \textbf{0.85} (21)
	&  0.80 
	&  0.70
	&0.47\\

\hline
\end{tabular}
\begin{flushleft}

Every cell contains mean value of AUC of the scoring function on 100 random split of the data on train and test in proportion 2:1. The cells for Smooth Rank contain in brackets  average number of features used.  The highest values in each row are marked by bold font.
\end{flushleft}
\end{table*}

Since SVM does not work with missing values, in all datasets, missing values were imputed using k-NN method with $k = 5.$

The Table 1 shows that Smooth Rank has the best results on 6 datasets, SVM (radial) - on 4 datasets, SVM (linear) - on 2 datasets, and RankBoost - on 5 datasets. However, RankBoost failed on the SPECTF; and on two other datasets  (Wpbc and Spine), it has average CI worse than the best method by 11 and 20 percents respectively, which can be considered a failure as well. Neither Smooth Rank nor SVM (radial) have such failures. Overall, average CI for each method  is: Smooth Rank - 0.8, SVM (radial) - 0.8,  SVM (linear) - 0.77, RankBoost - 0.75. 

The results indicate that Smooth Rank is not worse than any other  bipartite ranking method under comparison on these benchmarks. It works on par with SVM with radial kernels and better than two other methods. 



\section{Survival analysis as a bipartite ranking problem}

Survival analysis  deals with the  datasets, where each observation has three components:   covariate vector $x$, a positive survival time $t$ and an event indicator $\delta$, which  is equal to 1 if an event (failure) occurred, and zero if the observation is (right) censored at time $t.$ 

The prediction in survival analysis is generally understood as an estimate of an individual's risk of failure, but the concept of the risk is open for interpretation. The commonly accepted  criterion  of the risk modeling is Harrell's concordance index \cite{Tutorial} measuring agreement between the model's scores and the order of the failure times. The criterion is not directly related with any particular interpretation of the scores. Concordance index measures proportion of pairs of observations, where direction of difference in scores coincides with the direction of difference in times of failure. In absence of censored observations and ties, concordance index equals AUC. 

In the traditional approach associated with Sir David Cox \cite{Cox}, the research is concentrated on a time-dependent  ``hazard function''  $\Lambda(x, t)$:  event rate at time $t$ conditional on survival of the individual $x$ until time $t$ or later (that is, $T \ge  t$).  
 Cox proportional hazard (PH) regression, the most popular survival analysis method so far,  is based on the strong assumption that the hazard function has the form of $$\Lambda(x, t) = \lambda(t) \cdot  exp(\beta(x)),$$ where $\lambda(t)$ is  unknown time-dependent function, common for all individuals in the population. The result of the modeling is, actually,  not the individual time-dependent hazard functions $\Lambda(x, t)$, but rather these ``proportionality'' scores. 

As an alternative approach, we define the ``risk'' as a chance of having failure before certain time $\mathcal{T}$.   The advantage of this interpretation is that it is natural and common among medical practitioners.   In this case, the model's outcome indicates propensity of an individual to have an early failure. Splitting observations on ``early failure'' vs ``no early failure'' reduces the problem to bipartite ranking. 

Even though this interpretation  does not take into account time differences within each class on the training set, the models can be evaluated on the test data the same way, as in traditional survival analysis.  One can expect that the higher the risk of early failure, the earlier is actual failure time.  Therefore, the  concordance index between the model's scores and the failure times on the test data is a reasonable criterion of the modeling success for this interpretation as well.  This allows comparison in performance between the Smooth Rank and existing survival analysis methods.  

One may think that Smooth Rank is at disadvantage over survival analysis methods which use more information from the training data, including the actual times of the events and multidimensional relationship between the covariates and the outcome. However, our experiments show that discarding  unreliable, noisy information may, actually, benefit the generalization ability of the proposed method.  

\subsection{Application of Smooth Rank on survival analysis data}

Reduction of the survival analysis to bipartite ranking requires introduction of the threshold time $\mathcal{T}$, separating  the observations in two classes. Given the threshold $\mathcal{T}$, class of  ``early failure'' comprises only events which happened no later than $\mathcal{T}$. The class ``no early failure'' includes all observations with time beyond $\mathcal{T}$.  

According with this definition of the classes, the censored observations with time $t \le \mathcal{T}$ are excluded from classification, since their ``class'' is unknown. They do not participate in learning. 

Selecting the threshold $\mathcal{T}$ may be motivated by specifics of the applied problem. Here, the time threshold is  selected to balance the classes. Denote $L_T, H_T$ sizes of two classes, obtained with threshold $T$. Then, each threshold $T$ can be characterized by the difference $h_ T = |L_T - H_T|.$ The time $\mathcal{T}$ is calculated as a time of an event which minimizes the criterion $h_T$. 

Unlike {\it median survival} time, which may not be reached on some datasets, the class-balancing threshold $\mathcal{T}$ can be determined for every dataset, where there are events.  

\subsection{Methods for comparison in survival analysis}

Cox PH regression implemented in the function {\it coxph} from the R package {\it survival} is used in our experiments as a baseline method.  

Cox PH regression  has  important limitations on input data. It can not be used when number of features exceeds the number of instances.  But actual field of application of this method is even narrower. For example, the tutorial \cite{Tutorial}  suggests that the number of covariates should not exceed 1/10 of the number of non-censored observations. Most of  advanced methods for prediction in survival analysis are developed to make this traditional approach more robust against overfitting on sparse data (see surveys in \cite{Segal, Survey2}). Overall, the two major directions for improvements involve (1) preliminary feature aggregation to lower problem dimensionality  and (2) regularization of the Cox regression to increase method's robustness. The shortcoming of feature aggregation is that it may produce uninterpretable decisions. Among the regularization methods, $L_1$ -penalized Cox regression is the most attractive because it produces concise interpretable rules. Several authors   \cite{Segal, Survey2 , vanBelleSuperior } indicate that $L_1$ -penalized Cox regression has robust performance as well.  

It is why $L_1$ -penalized Cox regression was selected as a second approach for comparison with Smooth Rank. The method CoxPath 
\cite{Park}  builds $L_1$ -penalized Cox regression models for several values of the regularization parameter $\lambda$. The ``best'' model is selected.  Used here is the function {\it CoxPath} implemented in  in the R package \textit{glmpath} by the method's authors. For each model, the function outputs values of three criteria: AIC, BIC, loglik. The criterion AIC was chosen here to select the best model for the given training set. All other parameters of the \textit{CoxPath} procedure were taken as default.

\section{Experiments on survival analysis data}

The selected methods were applied on 10 survival analysis datasets. 

\begin{itemize}	
\item BMT:  The dataset represents data on 137 bone marrow transplant patients \cite{BMT} . The data allow to model several outcomes. Here,  the  models are built for disease free survival time. The first feature is diagnosis, which has three values:       ALL; AML Low Risk; AML High Risk. Other features characterize demographics of the patient and donor, hospital, time of waiting for transplant, and some characteristics of the treatment. There are 11 features overall, among them two are nominal. 

\item Colon: These are data from one of the first successful trials of adjuvant chemotherapy for colon cancer. Levamisole is a low-toxicity compound previously used to treat worm infestations in animals; 5-FU is a moderately toxic (as these things go) chemotherapy agent. There is possibility to model two outcome: recurrence and death. The data can be found in R package ÒsurvivalÓ.  The features include ÒtreatmentÓ (with three options: Observation, Levamisole, Levamisole+5-FU); properties of the tumor, number of lymph nodes. There are total 11 features and 929 observations. 

\item Lung1: Survival in patients with advanced lung cancer from the North Central Cancer Treatment Group \cite{Lung1}. Performance scores rate how well the patient can perform usual daily activities. Other features characterize calories  intake and weight loss. The dataset has 228 records with 7 features

\item  Lung2, the dataset from \cite{Lung2} Along with the patients' performance scores, the features include cell type (squamous, small cell, adeno, and large), type of treatment and prior treatment. 

\item BC : Breast cancer  dataset \cite{BC}. It contains 7 tumor characteristics in 97 records of patients. 

\item PBC : This data is from the Mayo Clinic trial in primary biliary cirrhosis of the liver conducted between 1974 and 1984 \cite{PBC}. Patients are characterized by standard description of the disease conditions.   The dataset has 17 features and 228 observations.

\item Al:  The data  \cite{Al} of the 40 patients with diffuse large
B-cell lymphoma contain information about 148 gene expressions associated with cell proliferation from ÒlympohichipÓ microarray data. Since there are more features than the observations, the Cox regression could not be applied on the data. The missing values were imputed using  the R program  {\it SVDImpute} from the package {\it imputation}

\item Ro02s: the dataset from \cite{Ro02} contains information about 240 patients with lymphoma. Using hierarchical cluster analysis on whole dataset and expert knowledge about factors associated with disease progression, the authors identified relevant four clusters and a single gene out of the 7399  genes on the lymphochip. Along with gene expressions, the data include two features for histological grouping of the patients. The authors aggregated gene expressions in each selected cluster to create a ÒsignaturesÓ of the clusters. The signatures, rather than gene expressions themselves were used for modeling.  The dataset with aggregated data has 7 features. 

\item Ro03g, Ro03s: the data \cite{Ro03} of 92 lymphoma patients. The input variables include data from ÒlymphochipÓ as well as results of some other tests. The Ro03s data contain averaged values of the gene expressions related with cell proliferation (proliferation signature). The Ro03g dataset includes the values of the gene expressions included in the proliferation cluster, instead of their average. Thus, the Ro03s dataset contains 6 features, and the dataset Ro03g contains 26 features. 
\end{itemize}

The results are presented in the Table 2. The ratio $N/M$ is included as a measure of the dataset ``sparsity'': generally, the smaller is the ratio, the less representative (more sparse) is the dataset.  

\begin{table*}
\caption{Comparison of Methods on Survival Analysis Data}
\centering
\begin{tabular}{| l | l | l | l | l|  l |  l  | l |}
\hline
$\#$ & Data & N $\times$ M &  N/M & Smooth Rank & Cox & Cox Path   \\ 
\hline
1 & BMT  &  137 $\times$ 11 & 12.4  
	& \textbf{0.68} (6.4)
	& 0.58  
	& 0.58 
	
	 \\

2 & Colon & 929 $\times$ 11 & 84.4 
	& \textbf{0.65} (4)
	& \textbf{0.65}  
	&  \textbf{0.65} 
	
	 \\
	
3 & Lung1 &  228 $\times$ 7 & 32.6 
	& \textbf{0.63} (5.7)
	& 0.62  
	& 0.62  
	 \\ 

4 & Lung2 &  137 $\times$ 6 & 22.8 
	       & \textbf{0.73} (2.16)
		   & 0.69   
	       & 0.70  
	       \\

5 & BCW &  97 $\times$ 7 & 13.9 
	& \textbf{0.71}  (5.9)
	& 0.69 
	& 0.69  
	\\
	
6 & PBC & 418 $\times$ 17 & 24.6 
	& \textbf{0.83} (12.6)
	& 0.82  
	& 0.82 

	\\
	
7 & Al & 40 $\times$ 148 & 0.27 
	& \textbf{0.63} (110) 
	& ---  
	& 0.52 

	\\
		
8 & Ro02s  & 240 $\times$ 7 & 34.3 
	& 0.70 (7)
	& \textbf{0.73}  
	& \textbf{0.73}  
	 \\
	
9 & Ro03s & 92 $\times$ 6 & 15.3 
	& \textbf{0.76} (3)
	& 0.74  
	& 0.75  
	\\
	
10 & Ro03g &  92 $\times$ 26 & 3.54 
	& \textbf{0.76} (23)
	& 0.58 
	& 0.67 
	\\

\hline
\end{tabular}

{Every cell contains mean value of CI on the test data for 100 random splits in proportion 2:1.  The highest values in each row are marked by bold font. The number in brackets is average number of features left after post-filtering in Smooth Rank}

\end{table*}

The Table 2 shows that in 9 out of 10 cases the Smooth Rank has the best results. 

In the three cases with the lowest ratio of $N/M$ (lines 1,7,10) the advantage of Smooth Rank is the most prominent. Its performance is higher than performance of  other methods by $9\% - 11\%.$ 

As an additional proof that Smooth Rank suffers less from the ``curse of dimensionality'', one can compare the results on the  datasets R003s, Ro03g which have information on the same patients with the same outcome. Dataset  Ro03g contains original values of gene expression, while Ro03s includes aggregated features, ``signatures''.  Smooth Rank has equally good (the best) results with or without aggregation, while other methods require preliminary feature aggregation for comparable performance.

The results indicate that proposed approach is a valid robust method of regularization, suited for particularly difficult and noisy data typical for survival analysis. Clear advantage of Smooth Rank on these 10 datasets justifies proposed reduction of survival analysis to bipartite ranking problem. 

For comparison, the regularization approach in CoxPath does not lead to significant performance improvement over Cox PH regression in most of cases. A possible explanation is that optimal model selection on the same small training set as part of the path method leads to ``model selection bias'' and limits advantage of the regularization with fixed parameter.

\section{Conclusions}

Sparse noisy datasets present especial challenge for learning. Common regularization approach with restriction on model weights may not be helpful, especially when it requires using the same sparse data for tuning method's parameters. Well known ``model selection bias''  increases the method's sensitivity to peculiarities of noisy data \cite{ModelSelectionBias}. Introduced here bipartite ranking method Smooth Rank represents an alternative approach to regularization. The approach includes (1) unsupervised aggregation of independently built predictors and (2) adjustment of hyper-parameters to the properties of the data by empiric formulas. The method was applied on two types of problems and compared with popular algorithms in each field.   The experiments demonstrated that Smooth Rank can be favorably compared with SVMs and RankBoost on two-class benchmark datasets. On survival analysis data, the same method showed significantly better generalization ability than Cox PH and CoxPath algorithm.  

\bibliography{survival}
\end{document}